\documentclass{article}

\usepackage{microtype}
\usepackage{graphicx}
\usepackage{subfigure}
\usepackage{booktabs} 

\usepackage{amsmath}
\usepackage{amssymb}
\usepackage{algorithm}
\usepackage{amsthm}
\usepackage{algorithmic}
\usepackage{array}
\usepackage{enumerate}
\usepackage{multirow}

\newtheorem{Theorem}{Theorem}
\newtheorem{Definition}{Definition}
\newtheorem{Lemma}{Lemma}

\usepackage{hyperref}

\usepackage[accepted]{icml2020}

\icmltitlerunning{On the Relation between Quality-Diversity Evaluation and Distribution-Fitting Goal in Text Generation}

\begin{document}

\twocolumn[
\icmltitle{On the Relation between Quality-Diversity Evaluation and Distribution-Fitting Goal in Text Generation}




\begin{icmlauthorlist}
\icmlauthor{Jianing Li}{ict,ucas}
\icmlauthor{Yanyan Lan}{ict,ucas}
\icmlauthor{Jiafeng Guo}{ict,ucas}
\icmlauthor{Xueqi Cheng}{ict,ucas}

\end{icmlauthorlist}

\icmlaffiliation{ict}{CAS Key Laboratory of Network Data Science and Technology, Institute of Computing Technology, Chinese Academy of Sciences, Beijing, China}
\icmlaffiliation{ucas}{University of Chinese Academy of Sciences, Beijing, China}

\icmlcorrespondingauthor{Yanyan Lan}{lanyanyan@ict.ac.cn}

\icmlkeywords{Text Generation, Evaluation Metric}

\vskip 0.3in
]



\printAffiliationsAndNotice{}  

\begin{abstract}
The goal of text generation models is to fit the underlying real probability distribution of text. For performance evaluation, quality and diversity metrics are usually applied. However, it is still not clear to what extend can the quality-diversity evaluation reflect the distribution-fitting goal. In this paper, we try to reveal such relation in a theoretical approach. We prove that under certain conditions, a linear combination of quality and diversity constitutes a divergence metric between the generated distribution and the real distribution. We also show that the commonly used BLEU/Self-BLEU metric pair fails to match any divergence metric, thus propose CR/NRR as a substitute for quality/diversity metric pair.
\end{abstract}

\section{Introduction}

Text generation is an essential task for many NLP applications, such as machine writing \citep{zhang2017flexible}, machine translation \citep{bahdanau2014neural}, image captioning \citep{rennie2017self} and dialogue system \citep{li2017adversarial}. Text generation models work by either explicitly modeling the probability distribution of text \citep{mikolov2010recurrent, yu2017seqgan}, or implicitly learning a generator which maps noise data to text \citep{zhang2017adversarial, chen2018adversarial}. Both approaches aim at generating text with the same distribution of given text data.

To achieve the distribution-fitting goal, divergence metrics are usually applied as the training objective for text generation models, which take minimal value 0 if and only if the model distribution exactly recover the real text distribution. Typical choices include the Kullback-Leibler divergence by maximum likelihood estimation (MLE) \citep{mikolov2010recurrent}, and Jensen-Shannon divergence or Wasserstein distance by adversarial training \citep{yu2017seqgan, gulrajani2017improved}. However during evaluation, divergence-based metrics fails to distinguish two under-fitting cases from each other: the low-quality case that generate unrealistic text, and the low-diversity case that generates dull and repeated text. As such, quality and diversity metrics are introduces to help the model diagnosis, such as BLEU \citep{papineni2002bleu} and Self-BLEU \citep{zhu2018texygen}. High generation quality requires the model to generate realistic samples, i.e. generated samples are free of grammatical or logical errors. High generation diversity requires the model to generate diverse samples, i.e. generated samples are less likely to be duplicate and contain diverse unique patterns.

Despite popular application of quality-diversity metrics in evaluation of text generation models \citep{chen2018adversarial, lu2018neural, fedus2018maskgan, alihosseini2019jointly}, the relationship between such evaluation and the distribution-fitting goal is still not clear. It seems to be a tacit consensus in recent works that a model with both higher quality and higher diversity also better fit the real text distribution \citep{caccia2018language, li2019differentiated, d2019training}. However, such assumption is yet to be verified. This is critical since a potential inequivalence may result in misleading evaluation conclusions. In this paper, we try to answer this question under the unconditional text generation setting by a theoretical approach.

To bridge the gap between distribution-fitting goal and quality-diversity evaluation, we require the optimal solutions from divergence minimization to be consistent with that of quality-diversity maximization. As such, we first give a general definition of quality and diversity. Then, we study a Multi-Objective Programming (MOP) problem which maximizes quality and diversity simultaneously. We prove there exists a family of Pareto-optimal solutions for this MOP problem, i.e. solutions which cannot be outperformed in terms of both quality and diversity. Then we prove the real distribution belongs to this Pareto-optimal family if and only if quality-diversity metrics are used in pairs with strong restrictions. Under such condition, a linear combination of quality and diversity constitutes a divergence metric between the generated distribution and the real distribution.

For quality-diversity metrics used in practice, we show that the widely applied BLEU/Self-BLEU metric pair fails to match any divergence metric. This is highlighted by a counter-intuitive observation that real text samples are significantly outperformed by manually constructed models over both BLEU and Self-BLEU. Therefore, we further propose Coverage Rate (CR) and Negative Repetition Rate (NRR) as substitute based on above theoretical analysis. Experiments show that CR/NRR act well as quality/diveristy metrics respectively, while a linear combination of CR/NRR acts well as divergence metric.

\section{Related Work}

To evaluate the performance of text generation models, many evaluation metrics are designed from different perspectives. Early neural text generation models use Perplexity (PPL) to show how well a language model fit the training data \citep{mikolov2010recurrent}. This is a divergence-based metric, and is still adopted in recent works \citep{fedus2018maskgan, lu2018cot, subramanian2018towards}. Calculation of PPL may be intractable for implicit models, so other divergence-based metrics are also practical choices, such as Kernel Density Estimation \citep{zhang2017adversarial}, Word Mover Distance \citep{lu2018cot}, MS-Jaccard \citep{alihosseini2019jointly}, and Frechet Distance \citep{semeniuta2018accurate, alihosseini2019jointly, d2019training}. However, divergence metrics provide limited information for model diagnosis, and may not correlate well with task performance \citep{chen1998evaluation, fedus2018maskgan}. Therefore, the quality and diversity of generated text are further considered as complementary metrics, which are also practical requirements in real applications \citep{zhang2018tailored, hashimoto2019unifying, gao2019jointly}.

For quality metrics, the evaluation is closely related to the ground truth distribution. \citet{yu2017seqgan} propose to use Negative Log-Likelihood where the real distribution is known in advance, which measures the average log-probability of generated samples over the real distribution. If the real distribution is not explicitly given, BLEU \citep{papineni2002bleu} and ROUGE \citep{lin2004looking} are usually applied, which measure the $n$-gram overlap between generated samples and a set of reference ground truth samples. For diversity metrics, the evaluation is performed within the model itself. \citet{li2015diversity} proposed Distinct-$n$ as diversity metric, which calculates the ratio of unique $n$-grams in generated samples. \citet{zhu2018texygen} proposed Self-BLEU, which is similar to BLEU but use generated samples as reference set.

There was a time in the past that only quality metrics are applied for evaluation, such as in works of SeqGAN \citep{yu2017seqgan}, RankGAN \citep{lin2017adversarial}, and LeakGAN \citep{guo2017long}. However after an observation of the quality-diversity tradeoff problem, \citet{zhu2018texygen} suggest to use a hybrid of both quality and diversity metrics, such as BLEU and Self-BLEU. This suggestion is widely adopted by many analytical works \cite{lu2018neural, caccia2018language, semeniuta2018accurate, alihosseini2019jointly}, as well as newly proposed methods, such as FM-GAN \citep{chen2018adversarial}, DDR \citep{li2019differentiated}, and ScratchGAN \citep{d2019training}. Despite the prevailing application of quality-diversity evaluation, its relationship with divergence metrics remains unclear, which poses great uncertainty for evaluation conclusions. Our work will help to build bridges between quality-diversity and divergence, and provide guidance for choosing appropriate quality-diversity metrics.

\section{Definition of Quality and Diversity}
Currently there is no unified definition for quality and diversity in text generation, which brings great challenges for further theoretical studies. In fact, it is not easy to define a general form of quality and diversity due to various understandings of these two aspects. Thus before moving on to further analysis, we first try to give a general form of quality and diversity in a mathematical view, though it may not be comprehensive enough to cover all possible understandings.

\subsection{A General Form of Quality and Diversity}
\label{section-form}
Text data is usually discrete, so we make the following notations. Assume the vocabulary size is $|V|$, and the maximum length is $L$, then the distribution of text data can be described by a categorical distribution with size $N=|V|^L$. We denote the real distribution and the generated model distribution as $P(x)=(P_1,P_2,\cdots,P_N)$ and $Q(x)=(Q_1,Q_2,\cdots,Q_N)$, respectively.

In general, the \textit{Quality} of a text generation model measures how likely the generated text are to be realistic text in human's view. Since the value of real probability $P(x)$ can be viewed as reflecting the realistic degree of a text $x$, the expectation of some function over $P(x)$ could be used to quantify quality. For example, in works of \citet{yu2017seqgan} and \citet{nie2018relgan}, \textit{Log-Likelihood (LL)} is used as the quality metric, where $LL(Q;P)=\mathbb{E}_{x\sim Q} \log P(x)$. Following this idea, we propose a general form of quality, i.e., $U(Q;P)=\mathbb{E}_{x\sim Q} f_u[ P(x)]$, where $f_u$ is a function over $P(x)$.

Similarly, the \textit{Diversity} of a text generation model measures how much difference there are among generated texts. From the viewpoint of information, \textit{Shannon-Entropy (SE)} of $Q(x)$ can be used as a natural diversity metric, where $SE(Q)=-\mathbb{E}_{x\sim Q} \log Q(x)$. From another understanding view, a text $x$ should be less likely to be generated again if the diversity is high. This idea has been adopted in biology to evaluate the diversity of biocoenosis, named as the \textit{Simpson's Diversity Index (SDI)}, where $SDI(Q)=1-\mathbb{E}_{x\sim Q} Q(x)$. Summarizing these two different understandings, we obtain a general form of diversity, i.e.~$V(Q)=-\mathbb{E}_{x\sim Q} f_v[Q(x)]$.

To this end, we propose a general form of quality and diversity metrics as follows:
\begin{equation*}
\begin{aligned}
U(Q) &= U(Q;P) = \mathbb{E}_{x\sim Q} f_u[P(x)] = \sum_{i=1}^N Q_i\cdot f(P_i), \\
V(Q) &= -\mathbb{E}_{x\sim Q} f_v[Q(x)] = \sum_{i=1}^N g(Q_i),
\end{aligned}
\end{equation*}
where $f_u(x)$ is denoted as $f(x)$ and $-x\cdot f_v(x)$ as $g(x)$.

\subsection{The Rationality of Quality and Diversity}
\label{section-rationality}
To guarantee $U$ and $V$ are rational quality and diversity metrics, we need to discuss about the conditions of $f$ and $g$. Without loss of generality, we first assume that $f$ is differentiable and $g$ is twice differentiable. Further, the following requirements are necessary for rational quality and diversity:
\begin{enumerate}
\item Generating more samples with higher real probability yields higher overall quality;
\item Distributing the probability more equally yields higher overall diversity.
\end{enumerate}
Mathematically, these two requirements can be formalized as the following two properties:

1. If $P_i>P_j$, then for $Q'=(Q_1, \dots, Q_i+\epsilon, \dots, Q_j-\epsilon, \dots)$, there is $U(Q')>U(Q)$ for any $\epsilon\in(0, Q_j)$.

2. If $Q_i\geq Q_j$, then for $Q'=(Q_1, \dots, Q_i+\epsilon, \dots, Q_j-\epsilon, \dots)$, there is $V(Q')<V(Q)$ for any $\epsilon\in(0, Q_j)$.

Then we can obtain the conditions of $f$ and $g$ by the following theorem:

\begin{Theorem}
\label{theorem-constraints}
The following conditions are both sufficient and necessary to satisfy the properties 1-2:
For any $x_1, x_2$ s.t. $x_1>x_2>0$ and $x_1+x_2\leq 1$, we have $f(x_1)>f(x_2)$ and $g'(x_1)<g'(x_2)$.
\end{Theorem}

According to Theorem \ref{theorem-constraints}, it is necessary for $f(x)$ to be strictly monotonically increasing and $g(x)$ to be strictly concave for $x\in(0,\frac{1}{2})$. For simplicity, we only consider the cases where such properties hold for $x\in(0,1)$, thus get a sufficient condition:
\begin{enumerate}
\item $f(x)$ is strictly monotonically increasing for $x\in(0,1)$;
\item $g(x)$ is strictly concave for $x\in(0,1)$.
\end{enumerate}

Under this condition, we can see that a model with highest quality will distribute all its density to text with highest real probability, and a model with highest diversity will be uniform, which are consistent with human understandings.

\section{Analysis of Quality-Diversity Evaluation}
\label{section-pareto}

In this section, we show how and to what extent can the quality-diversity evaluation reflect the distribution-fitting goal. The key idea is to solve the Multi-Objective Programming (MOP) problem which tries to maximize quality and diversity simultaneously. We give the structure of all the Pareto-optima of this MOP problem, which constitutes the Pareto-frontier. Then we prove the ground truth distribution lies in this frontier if and only if $f$ and $g$ are paired according to a given rule. Under such condition, a linear combination of quality and diversity constitutes a divergence metric, which means the quality-diversity evaluation is sufficient to reflect the distribution-fitting goal.

\subsection{The MOP Problem}
We consider the following MOP problem:
\begin{equation*}
\begin{aligned}
\max_Q\ (U(Q), &V(Q)) \\
s.t. \sum_{i=1}^N Q_i&=1 \\
\forall i,\ Q_i&\geq 0
\end{aligned}
\end{equation*}
The goal is to maximize both quality and diversity, while keeping $Q$ a legal distribution. The optimal solutions of a MOP problem are called Pareto-optima, which means no other solution can beat them consistently over all objectives.

We give definitions of the terminologies of Pareto-optimality below:

\begin{Definition}
\label{definition-dominate}
For two distributions $Q$ and $Q'$, if one of the following conditions are satisfied, we say that $Q$ is dominated by $Q'$.
\begin{enumerate}
\item $U(Q')>U(Q)$ and $V(Q')\geq V(Q)$;
\item $U(Q')\geq U(Q)$ and $V(Q')>V(Q)$.
\end{enumerate}
A solution $Q$ is called a Pareto-optimum if it is not dominated by any $Q'$. The set containing all the Pareto-optima is called the Pareto-frontier.
\end{Definition}

Intuitively, a Pareto-optimum is a solution that there is no distribution can achieve both higher quality and higher diversity than it. And all the Pareto-optima constitutes the Pareto-frontier. The Pareto-frontier may collapse into one solution which leads to a global optimum, e.g. if $P$ is uniform, the unique optimal solution would be $Q^*=P$. However it is often the case where the objectives in MOP problem cannot reach their optima consistently, which results in a family of optimal solutions. Therefore, the structure of the Pareto-frontier under a non-uniform $P$ is what we care about.

\subsection{The Pareto-frontier}

\begin{figure}[]
\centering
\includegraphics[width=0.40\textwidth]{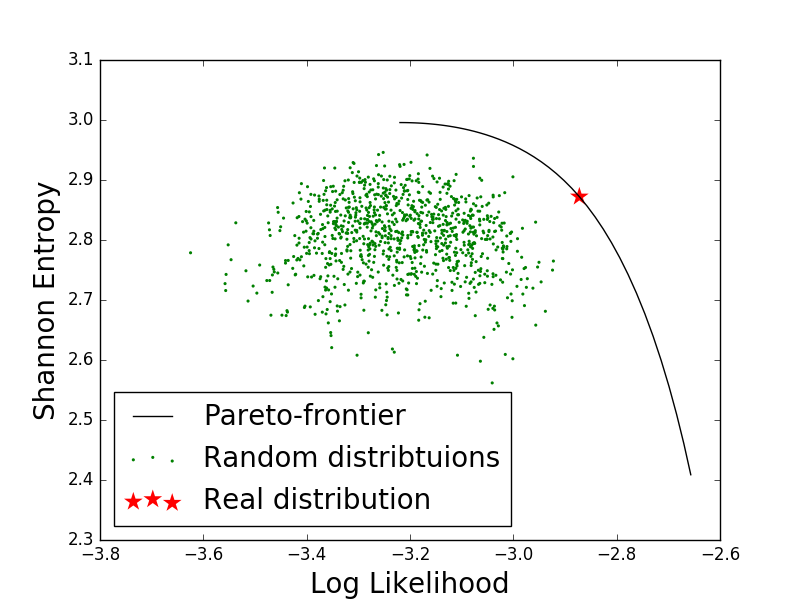}
\caption{Illustration of the Pareto-frontier of LL-SE metric pair on a toy categorical distribution, which contains $20$ categories and probabilities are sampled from uniform distribution with normalization.}
\label{figure-frontier}
\end{figure}

We show the structure of the Pareto-frontier by giving the following theorem:

\begin{Theorem}
\label{theorem-frontier}
For a distribution $Q$, if $P$ is not uniform, then:

(1) The following condition is both sufficient and necessary for $Q$ to be a Pareto-optimum: there exist real value $w\leq 0$ and $b$ that for any $i=1,\dots,N$, there is
\begin{equation*}
Q_i=\hat g'^{-1}[w\cdot f(P_i)+b],
\end{equation*}
where
\begin{displaymath}
\hat g'^{-1}(x) = \left\{ \begin{array}{ll}
g'^{-1}(x) & \textrm{if $x< g'(0)$,}\\
0 & \textrm{if $x\geq g'(0)$,}
\end{array} \right.
\end{displaymath}

(2) $b$ is correspondent to $w$, i.e. $b$ is fixed once $w$ is fixed. If $f(x)<0$ for all $x\in [0,1]$, then $b$ is strictly monotonically increasing w.r.t. $w$. If $f(x)>0$ for all $x\in [0,1]$, then $b$ is strictly monotonically decreasing w.r.t. $w$.

(3) Denote a Pareto-optimum $Q$ as $Q(w)$, then for any $w_1 < w_2$: if $w_1, w_2\in [B, 0]$, there is $Q(w_1)\neq Q(w_2)$ and $U(Q(w_1))>U(Q(w_2)), V(Q(w_1))<V(Q(w_2))$; if $w_1, w_2\in (-\infty, B]$, there is $Q(w_1)= Q(w_2)$; where $B=\frac{g'(\frac{1}{M})-g'(0)}{f(P_{m_1})-f(P_{m_2})}$, and $P_{m_1}=\max_i P_i$, $P_{m_2}=\max_{P_i\neq P_{m_1}} P_i$, $M=\#\{i|P_i=P_{m_1}\}$, \# denotes the cardinality of a set.
\end{Theorem}

According to Theorem \ref{theorem-frontier}, different $w$s lead to different distributions, so we can change $w$ from $0$ to $B$ and get a family of optimal solutions with different quality and diversity. As such, for a non-uniform $P$, the Pareto-frontier is a family of distributions.

We can see quality and diversity act as a tradeoff if we want to maximize them at the same time. Since all distributions in the Pareto-frontier are Pareto-optima, trying to improve one metric for an optimum will lead to another optimum at most, thus inevitably causing another metric to drop. This result provides support for the quality-diversity tradeoff problem observed in previous works \citep{zhu2018texygen, caccia2018language}.

We show the result of Theorem \ref{theorem-frontier} here on a special case. We pair Log-Likelihood (LL) with Shannon-Entropy (SE), the corresponding Pareto-optima can be written as
\begin{equation*}
Q_i=\frac{P_i^\beta}{Z},\ Z=\sum_{i=1}^N P_i^\beta, \ \beta\geq 0,
\end{equation*}
we have $w=-\beta$, and $b=1+\log Z$. These Pareto-optima are formerly used as quality-diversity tradeoff solutions by \citet{li2019differentiated}.

An illustration of the Pareto-frontier on a toy distribution is shown in Figure \ref{figure-frontier}. We can see that quality and diversity are negatively correlated for solutions in the Pareto-frontier. Note that the ground truth distribution lies exactly on the frontier in this LL-SE case, which can be checked by setting $\beta=1$. We will then show this is the key to the relation between quality-diversity metrics and divergence metrics.

\subsection{Relationship with Divergence}
\label{section-relationship}

To bridge the gap between the distribution-fitting goal and quality-diversity evaluation, it is necessary for the optimal solutions from divergence minimization to be consistent with that from quality-diversity maximization. Since $Q=P$ is the optimal solution with minimum divergence and the above Pareto-frontier is the set of optimal solutions with maximal quality and diversity, we require $Q=P$ to be in the Pareto-frontier. Theoretical results are shown in the following Theorem:

\begin{Theorem}
\label{theorem-groundtruth}
The following condition is both sufficient and necessary for $Q=P$ to be a Pareto-optimum for any $P$: there exist $w_0\leq 0$ and $b_0$ that
\begin{equation*}
g(x)=w_0\int_0^x f(u)\mathrm{d}u+b_0x.
\end{equation*}
If the above condition is satisfied, then $Q=P$ corresponds to a Pareto-optimum with $w=w_0$ and $b=b_0$, and it is the only distribution that maximize $\Psi(Q)=\alpha U(Q)+(1-\alpha)V(Q)$ with $\alpha=\frac{w_0}{w_0-1}\in [0,1)$, and $D(P||Q)=\Psi(P)-\Psi(Q)$ becomes a divergence metric.
\end{Theorem}

We find that if quality and diversity metrics are carefully chosen, namely $g$ is the integral of an affine transformation of $f$, we can get a divergence metric by a linear combination of these two metrics.

The LL-SE case satisfies the condition in Theorem \ref{theorem-groundtruth}. Under this special case, there is $\Psi(Q)=\frac{1}{2}\mathrm{LL}(Q)+\frac{1}{2}\mathrm{SE}(Q)$, and
\begin{equation*}
D(P||Q)=\frac{1}{2}\sum_{i=1}^N Q_i\cdot\log \frac{Q_i}{P_i},
\end{equation*}
which is exactly the Reverse KL divergence if the constant $\frac{1}{2}$ is ignored. This linearly combined divergence metric can be viewed as a tangent line of the Pareto-frontier curve in Figure \ref{figure-frontier}, and the real distribution is the tangent point.

Since such condition is also necessary, the real distribution is unlikely to be a Pareto-optima if we use casually chosen metrics. This means, there would be one distribution achieving both higher quality and higher diversity than the ground truth, which is implausible. Therefore, if the condition in Theorem \ref{theorem-groundtruth} is not satisfied, it would be unlikely to measure the divergence using a combination of quality and diversity.

Now we can conclude that, it is sufficient to reflect the distribution-fitting goal by a hybrid of quality-diversity evaluation. However, specific metrics should be chosen carefully, in order to avoid the potential violation of such property. Suppose such property is violated severely, featured by a huge gap between the ground truth distribution and the Pareto-frontier, then a model which perfectly fits the real distribution would be significantly outperformed by another model over both quality and diversity, resulting in misleading conclusions.

Therefore in the next section, we will examine the existence of the gap for quality-diversity metrics used in practice, and provide suggestions on the choice of quality-diversity metrics.

\section{Options for Quality-Diversity Metrics}

It is yet to be examined that whether existing quality-diversity metrics are sufficient to reflect the distribution-fitting goal. For metrics satisfying our defined general form in Section \ref{section-form}, conclusions can be drawn directly by applying Theorem \ref{theorem-groundtruth}. For example, the Log-likelihood (LL) is widely used as quality metric, which is correspondent to NLL-oracle \citep{yu2017seqgan} and Reverse PPL \citep{subramanian2018towards}. As proved above, LL satisfies the condition in Theorem \ref{theorem-groundtruth} if it's paired with Shannon Entropy (SE). Consequently, it is safe to use LL-SE together as in the work of \citet{alihosseini2019jointly}.

However for most scenarios with real text data, the calculation is intractable for the general form of quality-diversity in Section \ref{section-form} as the ground truth distribution is unknown, including the LL-SE pair. Practical metrics (e.g. BLEU and Self-BLEU) thus usually fall out of this framework, and Theorem \ref{theorem-groundtruth} cannot be applied directly. In order to make a judgement on such metrics, we suggest to consider the compatibility between divergence and quality-diversity metric pair. We say a pair of quality-diversity metrics is \textit{divergence-compatible} if the real distribution is a Pareto-optimum under the MOP problem maximizing both metrics. Such compatibility is a necessary condition for the existence of a corresponding divergence metric which is strictly monotonically decreasing w.r.t. both quality and diversity.

\subsection{BLEU and Self-BLEU}

BLEU \citep{papineni2002bleu} and Self-BLEU \citep{zhu2018texygen} are common metrics for quality and diversity evaluation, respectively. Intuitively, BLEU measures the $n$-gram overlap between a candidate set of generated text and a reference set of real text, while Self-BLEU is the average BLEU score of each generated text with other candidates as reference. High BLEU score means that $n$-grams in generated text are more likely to appear in real text, thus BLEU can be used as quality metric. Similarly, high Self-BLEU score means that generated text are similar to each other in terms of $n$-gram, thus Negative Self-BLEU (NSBLEU as abbreviation) can be used as diversity metric.

The expression of BLEU on a candidate set $C$ is:
\begin{equation*}
\begin{aligned}
&\mathrm{BLEU} = BP\cdot exp(\frac{1}{M}\sum_{n=1}^M \log p_n), \\
&p_n = \frac{\sum_{c\in C}\sum_{gram_n\in c} Count_{clip}(gram_n)}{\sum_{c'\in C}\sum_{gram'_n\in c'} Count(gram'_n)},
\end{aligned}
\end{equation*}
where $BP$ is the Brevity Penalty which penalizes short sentences, and $M$ denotes the maximum $n$-gram order. $p_n$ is a precision term, which measures the proportion of grams in the candidate set that also appear in the reference set. BLEU is the geometric mean of $p_n$ for all $n\leq M$, multiplied by a penalty term.

The expression of BLEU does not seem to satisfy the general form of quality/diversity defined in Section \ref{section-form}. However on some special case, the general form is still satisfied, upon which we show some symptoms indicating the incompatibility of BLEU-NSBLEU. Assume the lengths of text are all $1$, so that $M=1$ and $BP\equiv 1$. In this case, BLEU contains only one term, i.e. $\mathrm{BLEU}=p_1$. Then for candidate set $C$ and reference set $R$, the expectation of BLEU and NSBLEU over generated distribution $Q$ and real distribution $P$ would be
\begin{equation*}
\begin{aligned}
\mathop{\mathbb{E}}\limits_{C\sim Q,R\sim P}\ \mathrm{BLEU}(C, R) = \sum_{i=1}^N Q_i\cdot [1-(1-P_i)^{|R|}], \\
\mathop{\mathbb{E}}\limits_{C\sim Q}\ \mathrm{NSBLEU}(C) = - \sum_{i=1}^N Q_i\cdot [1-(1-Q_i)^{|C|-1}].
\end{aligned}
\end{equation*}
Such expressions satisfy the general form with
\begin{equation*}
\begin{aligned}
f(x)=1-(1-x)^{|R|}, \quad g(x)=-x+x\cdot(1-x)^{|C|-1}.
\end{aligned}
\end{equation*}
The condition in Theorem \ref{theorem-groundtruth} would be satisfied if and only if $|R|=1$ and $|C|=2$, which becomes $f(x)=x$ and $g(x)=-x^2$. However, the size of reference set $|R|$ is usually far more than $1$, under which cases the BLEU-NSBLEU metric pair would be divergence-incompatible.

Though above analysis is done on a special case, such results imply a potential incompatibility for general BLEU-NSBLEU metric pairs. We will confirm this incompatibility by an empirical approach in Section \ref{section-experiment}.

\begin{table*}[t]
\caption{Lower-bound of QDisc and DRate w.r.t. BLEU-NSBLEU on synthetic data with different $\sigma$s.}
\label{table-synthetic}
\begin{center}

\begin{tabular}{l|cc|cc|cc}
\hline
\multirow{2}{*}{Metrics} &
\multicolumn{2}{c|}{$\sigma=0.5$} &
\multicolumn{2}{c|}{$\sigma=1.0$} &
\multicolumn{2}{c}{$\sigma=2.0$} \\
\cline{2-7}
 & QDisc & DRate(\%) & QDisc & DRate(\%) & QDisc & DRate(\%) \\
\hline
BS-1 & 0.01287 & 2.55 & 0.01509 & 3.29 & 0.01063 & 3.15 \\
BS-2 & 0.02384 & 9.41 & 0.01699 & 4.27 & 0.01146 & 1.71 \\
BS-3 & 2.090 $\times 10^{-8}$ & $<$0.01 & 6.045 $\times 10^{-6}$ & 0.19 & 3.878 $\times 10^{-4}$ & 0.05 \\
\hline
\end{tabular}
\end{center}
\end{table*}

\subsection{The Proposed Metric Pair}

To avoid possible misleading conclusions in practice, we suggest to use diversity-compatible quality-diversity metric pair.

Since the real probability $P(x)$ is required in $U(Q;P)$ under the general form in Section \ref{section-form}, calculation of most quality metrics are intractable on real text data. The only exception is the case with $f(x)=x$, paired with $g(x)=-x^2$. The linearity of $f$ can avoid the explicit form of $P(x)$ by sampling from real data, i.e. $U(Q)=\mathbb{E}_{x\in P}\ Q(x)$. We name the corresponding quality metric as \textit{Coverage Rate (CR)}, and diversity metric as \textit{Negative Repetition Rate (NRR)}. Even so, we observe a large variance while estimating CR and NRR on real text data. This is mainly because of the extremely large space of text of $N=|V|^L$. Therefore, estimations of CR/NRR are highly inaccurate in the text space.

We thus suggest to calculate CR-NRR in $n$-gram space rather than in text space. Derive the $n$-gram distribution $Q_g$ and $P_g$ from text distribution $Q$ and $P$, so that
\begin{equation*}
\begin{aligned}
\mathrm{CR}_n(Q; P) &= \sum_{gram_n\in S_n} Q_g(gram_n)\cdot P_g(gram_n), \\
\mathrm{NRR}_n(Q) &= - \sum_{gram_n\in S_n} Q_g^2(gram_n),
\end{aligned}
\end{equation*}
where $S_n$ denotes the set of all possible $n$-grams. In practice, $Q_g$ and $P_g$ can be estimated by the empirical distribution, i.e. count the number of target $n$-grams and divide by the total number. Note that if calculated by the longest $n$-gram with $n=L$, $\mathrm{CR}_n$ and $\mathrm{NRR}_n$ would exactly recover the original CR and NRR metric in text space, thus can be viewed as a generalized form. In the rest of this paper, we use \textit{CR-NRR} as a default notation in the $n$-gram space unless explicitly stated.

In the $n$-grams space, calculation of metric pairs with other $f$/$g$ functions also becomes possible. However, metrics such as LL-SE suffer from another smoothing problem on real text data, i.e. their values go to infinity if some $n$-grams do not appear in candidate set or reference set. Therefore, we still suggest to use CR-NRR as a first choice.

Though there is a conversion from the text space to the $n$-gram space, CR/NRR can still reflect quality/diversity. The $\mathrm{CR}_n$ metric measures the average probability for an $n$-gram in candidate set to appear in the reference set, thus is an indicator of quality. Similarly, $\mathrm{NRR}_n$ measures the average probability for an $n$-gram to appear again in two consecutive sampling processes over the candidate set, thus is an indicator of diversity.

We then check the divergence-compatibility of CR-NRR evaluation. Firstly, CR-NRR is divergence-compatible w.r.t. distributions in the $n$-gram space, according to Theorem \ref{theorem-groundtruth}. We name the corresponding divergence metric as \textit{CR-NRR Divergence (CND)}, where
\begin{equation*}
\Psi_n(Q)=\frac{2}{3}\mathrm{CR}_n(Q; P)+\frac{1}{3}\mathrm{NRR}_n(Q),
\end{equation*}
and
\begin{equation*}
\begin{aligned}
\mathrm{CND}_n(Q; P) &= 3\cdot[\Psi_n(P) - \Psi_n(Q)] \\
=&\sum_{gram_n\in S_n} [Q_g(gram_n)- P_g(gram_n)]^2.
\end{aligned}
\end{equation*}
Secondly, CR-NRR is also divergence-compatible w.r.t. distributions in the text space. Assume $Q=P$ is dominated by $Q'$ under CR-NRR evaluation, which means $Q_g=P_g$ would also be dominated by $Q'_g$. This cause contradiction with the compatibility in $n$-gram space, so the compatibility in text space also holds.

In addition to the divergence-compatibility property, CR-NRR is also easy to acquire. It does not require the explicit value of $P(x)$ or $Q(x)$, thus can be applied on implicit models similarly to BLEU-NSBLEU. Moreover, the time complexity of CR-NRR algorithm is $O(m+n)$, which is much lower than BLEU-NSBLEU with $O(m\cdot (m+n))$, where $m$ and $n$ denote the size of candidate and reference set respectively. To conclude, we suggest to use CR-NRR in $n$-gram space for quality-diversity evaluation, instead of BLEU-NSBLEU.

\section{Experiments}
\label{section-experiment}

In this section, we perform compatibility analysis of BLEU-NSBLEU, compared with CR-NRR on both synthetic data and real text data. We show that BLEU-NSBLEU is significantly divergence-incompatible, by observing a phenomenon that ground truth text data are clearly outperformed over both BLEU and NSBLEU by some manually constructed model. We also show that CR/NRR are representative for quality/diversity evaluation respectively, while CND is representative for divergence evaluation.

To measure the degree of incompatibility, we calculate the Quality Discrepancy (QDisc) and Discrepancy Rate (DRate):
\begin{equation*}
\begin{aligned}
\mathrm{QDisc} &= \max_Q U(Q)-U(P),\ s.t.\ V(Q)\geq V(P), \\
\mathrm{DRate} &= \frac{\mathrm{QDisc}}{\max_Q U(Q)-U(Q')},\ Q'=\mathop{argmax}\limits_Q V(Q).
\end{aligned}
\end{equation*}
Intuitively, we try to find a model with best quality while its diversity is no lower than that of real distribution. Then QDisc measures the difference between this model and the real distribution in terms of quality. DRate measures the ratio between QDisc and the total range of quality for all Pareto-optima. A metric pair is divergence-compatible if and only if $\mathrm{QDisc}=0$.

\subsection{Experiments on Synthetic Data}

\begin{figure*}[]
\centering
\subfigure[MSCOCO dataset]{
\includegraphics[width=0.24\textwidth]{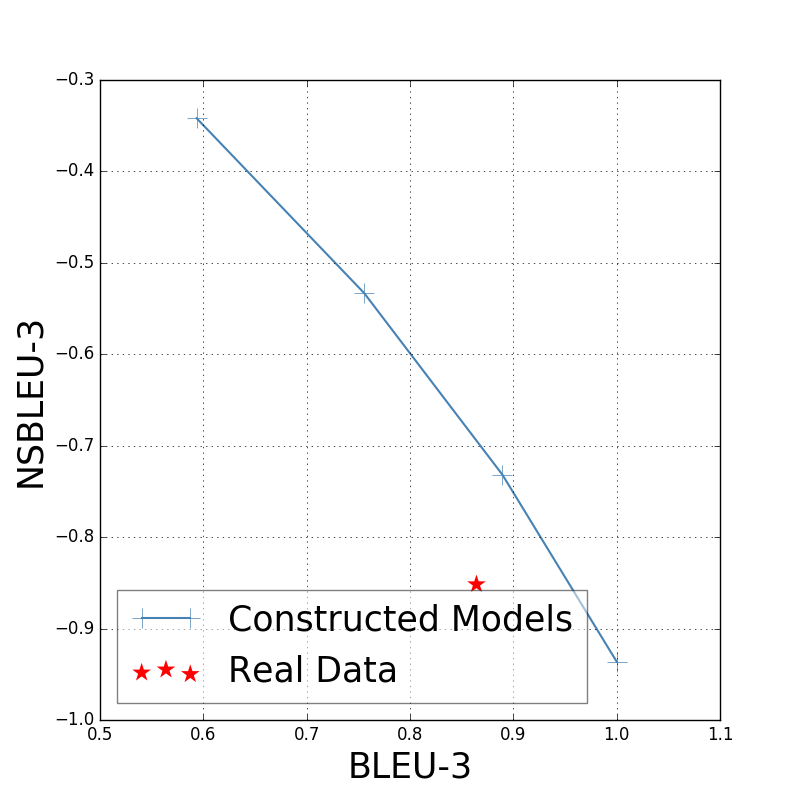}
\includegraphics[width=0.24\textwidth]{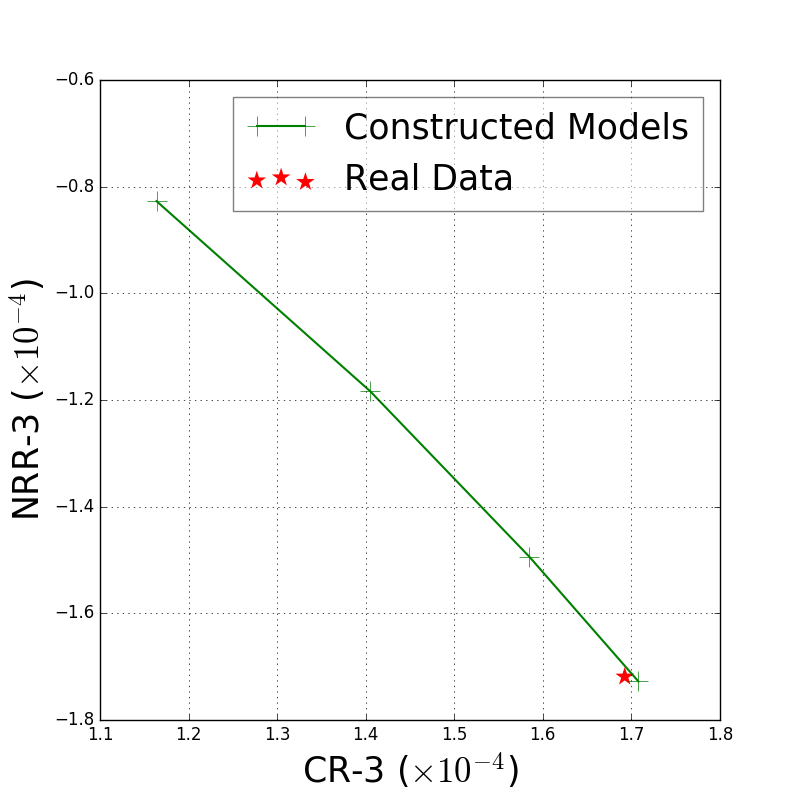}
}
\subfigure[WMT dataset]{
\includegraphics[width=0.24\textwidth]{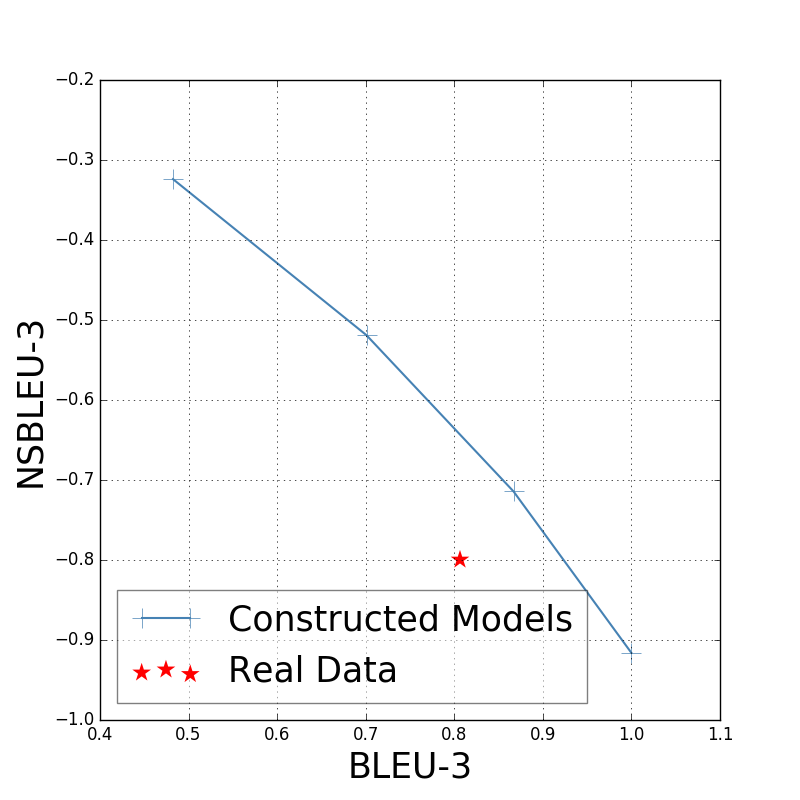}
\includegraphics[width=0.24\textwidth]{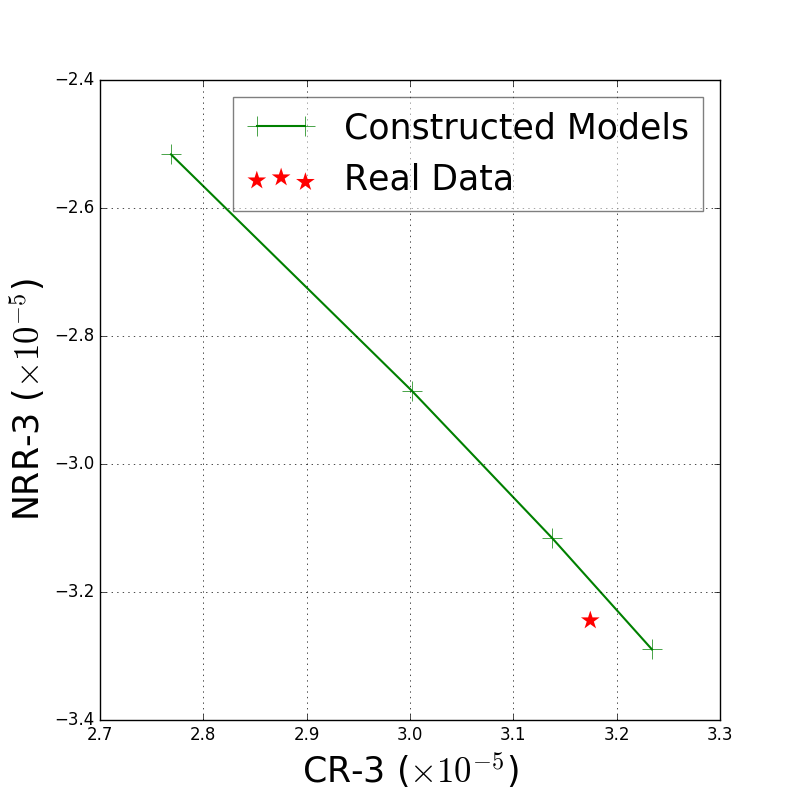}
}
\caption{Evaluation of BLEU-NSBLEU and CR-NRR on real text data. Test data are random text from reference set, mixed with noise with a proportion of $\epsilon=[0.0, 0.2, 0.4, 0.6]$ from right to left.}
\label{figure-realdata}
\end{figure*}

\begin{table*}[t]
\caption{Estimation of QDisc, DRate, Self-Ratio, and Ref-Ratio on real text data.}
\label{table-real}
\begin{center}

\begin{tabular}{l|cccc|cccc}
\hline
\multirow{2}{*}{Metrics} &
\multicolumn{4}{c|}{MSCOCO} &
\multicolumn{4}{c}{WMT} \\
\cline{2-9}
 & QDisc & DRate(\%) & Self-Ratio & Ref-Ratio & QDisc & DRate(\%) & Self-Ratio & Ref-Ratio\\
\hline
BS-2 & 0.032 & 3.2 & 0.034 & 0.314 & 0.034 & 3.4 & 0.036 & 0.26 \\
BS-3 & 0.090 & 9.0 & 0.104 & 0.814 & 0.117 & 11.7 & 0.145 & 0.88 \\
BS-4 & 0.162 & 16.2 & 0.219 & 1.46 & 0.211 & 21.1 & 0.339 & 1.59 \\
\hline
CN-2 & 0.75$\times 10^{-6}$ & 0.013 & 0.0005 & 0.006 & 3.69$\times 10^{-7}$ & 0.016 & 0.0008 & 0.025 \\
CN-3 & 1.07$\times 10^{-6}$ & 0.079 & 0.0063 & 0.087 & 3.45$\times 10^{-7}$ & 0.098 & 0.0109 & 0.358 \\
CN-4 & 1.15$\times 10^{-6}$ & 0.163 & 0.0247 & 0.421 & 3.12$\times 10^{-7}$ & 0.220 & 0.0525 & 2.092 \\
\hline
\end{tabular}
\end{center}
\end{table*}

We first run experiments on synthetic data rather than real text data, in order to get the precise values of all metrics. Under this setting, the information of generated distribution $Q$ and real distribution $P$ are explicitly given in advance, thus eliminates the possible variance from sampling. The synthetic data are texts with length $L$ using a pseudo vocabulary $V$. We construct the real distribution using an oracle LSTM model as in SeqGAN \citep{yu2017seqgan}, whose weights are randomly sampled from a gaussian distribution with $\mu=0$. Different standard deviation $\sigma$s are applied to get several synthetic real distributions with different levels of entropy, i.e. distribution with smaller $\sigma$ is more flat and of higher entropy, and distribution with larger $\sigma$ is more sharp and of lower entropy.

Calculation of QDisc and DRate can be achieved by a simple binary-search algorithm if the exact form of Pareto-frontier is known. However for BLEU-NSBLEU metric pair, the frontier is unknown since Theorem \ref{theorem-frontier} cannot be applied in this case. Consequently, we opt to used an optimization-based method for the estimation of QDisc. We try to solve the following optimization problem using stochastic gradient descent (SGD) with momentum:
\begin{equation*}
Q^* = \mathop{\mathrm{argmax}}\limits_Q\ U(Q) - \lambda\cdot \max(0, V(P)-V(Q)),
\end{equation*}
where $\lambda$ is a penalty term to discourage the case where divergence is lower than real distribution $P$. We set $\lambda=2.0$ in our experiments. So that $\mathrm{QDisc}=U(Q^*)-U(P)$, and the denominator in DRate is also calculated through such optimization-based method.

For BLEU metric with candidate set size $m$ and reference set size $n$, the expectation can be directly calculated by
\begin{equation*}
\begin{aligned}
&\mathop{\mathbb{E}}\limits_{C\sim Q,R\sim P}\ \mathrm{BLEU}(C, R) = \\
&\sum\limits_{C\in V^{L\cdot m}, R\in V^{L\cdot n}}\ \prod_{i=1}^m Q(C_i)\cdot \prod_{j=1}^n P(R_j)\cdot \mathrm{BLEU}(C, R).
\end{aligned}
\end{equation*}
The time complexity (number of terms) of such calculation is $O(|V|^{L\cdot (m+n)})$. This is intolerable for above optimization problem even in text space of normal size. As a result, we set $|V|=4, L=3, m=1, n=2$, and apply SGD under the Tensorflow framework\footnote{Slight increase of any parameter will consume intolerably more time, and is not necessary for the conclusions.}.

We use CN-n and BS-n as abbreviation for CR-NRR and BLEU-NSBLEU with $n$-gram, respectively. We report the QDisc and DRate of BLEU-NSBLEU in Table \ref{table-synthetic}. Note that the reported QDisc values are corresponding lower bounds, since the optimization-based method does not guarantee a global optimum. These non-zero QDisc values provide a clear support for the incompatibility of BLEU-NSBLEU. We can also see that such discrepancy is significant on some cases, e.g. $\mathrm{QDisc}>0.02$ and $\mathrm{DRate}=9.41\%$ for BS-2 on data with $\sigma=0.5$. A QDisc value of 0.02 means that, we cannot surely claim that a model is better than another when the quality gap is below 0.02, which is already a clear gap for BLEU. We also run similar experiments for CR-NRR. However, no positive lower bound is observed, which is in accordance with our theory.

\subsection{Experiments on Real Text Data}

Significance of quality discrepancy varies on different cases, thus we care about the discrepancies on real text data. We use two public datasets, MSCOCO Image Caption dataset \citep{chen2015microsoft} and EMNLP2017 WMT News dataset\footnote{http://statmt.org/wmt17/translation-task.html}.
We use 50,000 sentences as candidate set and another 50,000 as reference set for each dataset \footnote{See supplementary material for detailed configurations.}.

To provide an estimation of QDisc and DRate, we manually construct a family of strong models. We mix the empirical distribution $\tilde{P}$ with truncated uniform distribution $M$ under different proportions, i.e. $Q=(1-\epsilon)\cdot\tilde{P}+\epsilon\cdot M$. During text generation, a random text from reference set is sampled with probability $1-\epsilon$, otherwise a text with random tokens of length $L'$ is constructed with probability $\epsilon$. We try both $L'=5$ and $L'=L$, and report the case with larger QDisc value.

We estimate QDisc by a linear interpolation between two closest points on the curve w.r.t. quality of real data. For the denominator of DRate in BLEU-NSBLEU, we use $1.0$ directly, since $\mathrm{BLEU}=1$ is reached for highest quality with $\epsilon=0.0$, and $\mathrm{BLEU}\approx 0$ for highest diversity with $\epsilon=1.0$. For CR-NRR, CR goes to $0$ when diversity is maximized with $\epsilon=1.0$. As for the maximal value of CR, we estimate it by using a single reference sentence as candidate and select the one with maximal CR value.

For a clearer view of the significance of quality discrepancy, we introduce two additional metrics: Self-Ratio and Ref-Ratio. Self-Ratio calculates the ratio between QDisc and the quality of candidate set. Ref-Ratio calculates the ratio between QDisc and the quality difference of $\epsilon=0.0$ and $\epsilon=0.2$. The evaluation results of BLEU-NSBLEU and CR-NRR with $3$-gram under $L'=5$ are shown in Figure \ref{figure-realdata}.

We can see that real data stays close to the CR-NRR curve, while a much larger gap is observed between real data and the BLEU-NSBLEU curve. We give the values of QDisc, DRate, Self-Ratio, and Ref-Ratio in Table \ref{table-real}. BLEU-NSBLEU shows a significant incompatibility, by QDisc values ranging from 0.032 to 0.211. Such huge discrepancy in BLEU is unbearable in real applications, e.g. we cannot claim a model is better than another even if it achieves higher NSBLEU and significantly higher BLEU. As a result, we suggest not to use BLEU-NSBLEU in order to avoid misleading conclusions. CR-NRR also shows a small positive discrepancy, this is due to the inevitable difference between the empirical distributions of candidate set and reference set. However, discrepancy caused by such distribution difference is generally much smaller than BLEU-NSBLEU. We also observe that DRate grows quickly as $n$-gram becomes longer for CR-NRR, thus we suggest to use CR-NRR with short $n$-gram such as CN-2 or CN-3.

\begin{figure*}[]
\centering
\subfigure[MSCOCO dataset]{
\includegraphics[width=0.24\textwidth]{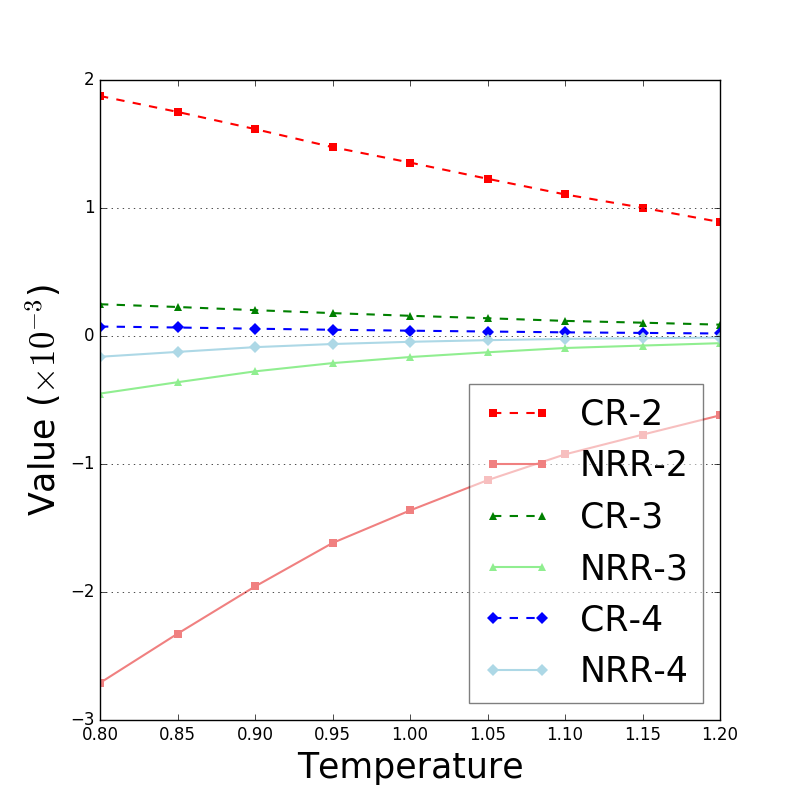}
\includegraphics[width=0.24\textwidth]{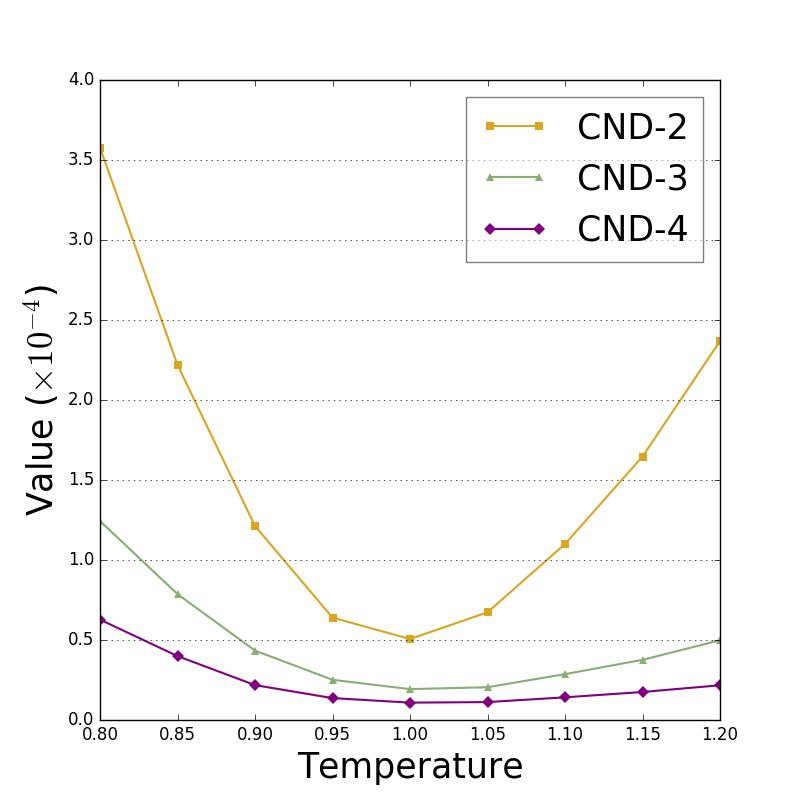}
}
\subfigure[WMT dataset]{
\includegraphics[width=0.24\textwidth]{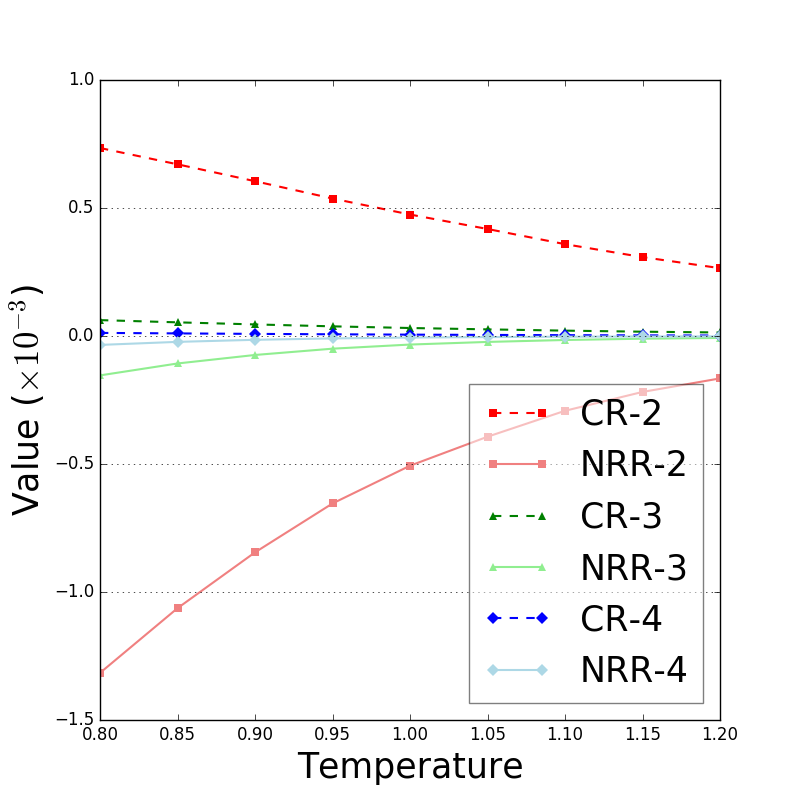}
\includegraphics[width=0.24\textwidth]{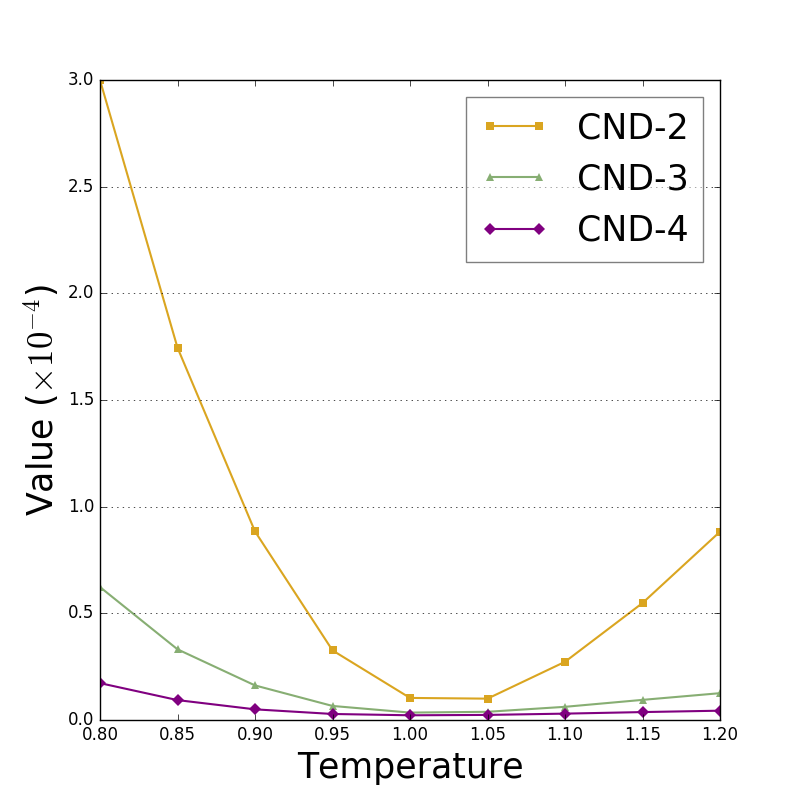}
}
\caption{Evaluation of CR-NRR and CND on real text data. Test data are generated by temperature-sweep on pre-trained RNNLMs.}
\label{figure-temperature}
\end{figure*}

Next we show how CR/NRR/CND behave on real text data. We apply temperature sweep on an RNN-based language model (RNNLM) pre-trained by maximum likelihood estimation, which is a quick way to get a family of models with quality-diversity tradeoff according to works of \citet{caccia2018language}. The RNNLM consists of an embedding layer, an LSTM layer, and a fully-connected output layer. The embedding dimension and number of hidden nodes are all set to 128. We train the model using Adam \citep{kingma2014adam} optimizer with learning rate $0.001$ by 30 epochs. As temperature $t$ grows, the model becomes more close to uniform, so that quality decreases and diversity increases, and minimal divergence is taken near $t=1.0$. Results are shown in Figure \ref{figure-temperature}, where we can see CR/NRR/CND are representative for quality/diversity/divergence respectively, which clearly fit our expectations. Therefore, we suggest to use CR-NRR for quality-diversity evaluation.

\section{Discussion}

Our above conclusions are mainly drawn under the unconditional text generation setting, however, quality-diversity evaluation is also getting great attentions under conditional text generation settings, such as dialogue system \cite{vijayakumar2016diverse}, machine translation \cite{shen2019mixture} and image captioning \cite{ippolito2019comparison}. In this section, we give a brief discussion about quality-diversity evaluation under conditional text generation settings.

Due to different formalization of quality and diversity metrics, our conclusions cannot be directly transferred to conditional text generation settings. Under these settings, the quality of text $x$ under condition $c$ is still defined as monotonically increasing w.r.t. the real conditional probability $P(x|c)$. So that the overall quality metric becomes the expectation of text quality over $x$ and $c$, which is the case for BLEU. Meanwhile, diversity metrics have two different understandings. One is defined as the average diversity of conditional model distribution $Q(x|c)$ under different $c$, such as Pairwise-BLEU \cite{shen2019mixture}. The other is define as the diversity of marginal model distribution $Q(x)=\sum_{c} P(c)Q(x|c)$, such as Distinct \cite{li2015diversity}. Formalization of both quality and diversity metrics depart from ours in Section \ref{section-form}, and may result in different conclusions, thus require further separate analysis. Though such analyses are not covered here, our work provides a paradigm for future theoretical analysis, including metric definition, Pareto-optimality analysis, and divergence-compatibility judgement.

Another difference lies in the point of view of task goal. While the goal of unconditional text generation is to design models that better fit the text distribution, in conditional text generation however, better human evaluation results are viewed as final goal in most cases. Therefore in these cases, the main focus would be designing metrics that better reflect human evaluation as well as designing training objectives that achieve better evaluation. It is also anticipated that whether human evaluation is compatible with divergence. We regard these as our future work.

\section{Conclusion}

In this paper, we give theoretical analysis of the relation between quality-diversity evaluation and distribution-fitting goal. We show that when using properly paired quality-diversity metrics, i.e. $g(x)$ is the integral of an affine transformation of $f(x)$, a linear combination of quality and diversity constitutes a divergence metric between the generated distribution and the real distribution.
For metrics used in practice, we show the commonly used BLEU and Self-BLEU metric pair fails to reflect the distribution-fitting goal. For a substitute, we suggest to use CR-NRR instead as quality-diversity metric pair.

\section*{Acknowledgement}
This work was supported by Beijing Academy of Artificial Intelligence (BAAI) under Grants No. BAAI2019ZD0306, and BAAI2020ZJ0303, the National Natural Science Foundation of China (NSFC) under Grants No. 61722211, 61773362, 61872338, 61902381, and 61906180, the Youth Innovation Promotion Association CAS under Grants No. 20144310, and 2016102, the National Key RD Program of China under Grants No. 2016QY02D0405, the Lenovo-CAS Joint Lab Youth Scientist Project.

\bibliography{icml2020}
\bibliographystyle{icml2020}

\appendix
\section*{Appendix}
\section{Preliminaries}
Before starting the proofs, we first introduce some preliminaries on the constrained convex optimization problem.
Assume $f(x)$, $c_i(x)$, and $h_j(x)$ are continuous differentiable function define on $\mathbb{R}^n$, consider the constrained convex optimization problem defined as follows:

\begin{equation}
\label{equation-ccop}
\begin{aligned}
\min_{x\in \mathbb{R}^n} \quad &f(x) \\
s.t. \quad &c_i(x)\leq 0,\quad i=1,2,\cdots,k \\
&h_j(x)=0,\quad j=1,2,\cdots,l
\end{aligned}
\end{equation}

The optimal solutions for above problem are given by the Lagrange Multiplier approach , as shown in the following theorem:

\begin{Theorem}
\label{theorem-kkt}
Assume $f(x)$ and $c_i(x)$ are convex, $h_j(x)$ are affine, and $c_i$ are strictly feasible (there exists one $x$ satisfying $c_i(x)<0$ for all $i$). Define the Lagrange function as:
\begin{equation*}
L(x,\alpha,\beta)=f(x)+\sum_{i=1}^k \alpha_i c_i(x)+\sum_{j=1}^l \beta_j h_j(x),
\end{equation*}
where $\alpha\geq 0$. Then the the following conditions are both sufficient and necessary for $x$ to be a solution in problem \ref{equation-ccop}.
\begin{equation}
\label{equation-kkt}
\begin{aligned}
\nabla_x L(x^*,\alpha^*,&\beta^*)=0 \\
\nabla_\alpha L(x^*,\alpha^*,&\beta^*)=0 \\
\nabla_\beta L(x^*,\alpha^*,&\beta^*)=0 \\
\alpha_i^* c_i(x^*)=0, \quad i&=1,2,\cdots, k \\
c_i(x^*)\leq 0, \quad i&=1,2,\cdots, k \\
\alpha_i^* \geq 0, \quad i&=1,2,\cdots, k \\
h_j(x^*)=0, \quad j&=1,2,\cdots, k
\end{aligned}
\end{equation}
\end{Theorem}
The conditions in Equation \ref{equation-kkt} are called the Karush-Kuhn-Tucker(KKT) conditions.

\section{Proof of Theorem $1$}
For property 1, from $U(Q')-U(Q)=\epsilon f(P_i)-\epsilon f(P_j)=\epsilon[f(P_i)- f(P_j)]>0$, we get $f(P_i)>f(P_j)$. We then get the conclusion by setting $x_1=P_i$ and $x_2=P_j$.

For property 2, $V(Q')-V(Q)=[g(Q_i+\epsilon)+g(Q_j-\epsilon)]-[g(Q_i)+g(Q_j)]<0$ is true for any $Q_i>Q_j$. Denote $C=Q_i+Q_j$ and $r(x)=g(x)+g(C-x)$, then we have $V(Q')-V(Q)=r(Q_i+\epsilon)-r(Q_i)<0$ for any $Q_i,\epsilon$. Since $0<Q_i<Q_i+\epsilon<1$, we need $r'(x)<0$ for $x\in (0,1)$. Then, since $r'(x)=g'(x)-g'(C-x)<0$ is true for any $0<C-x<x<1$. Set $x_1=C-x$ and $x_2=x$ and we get $g'(x_1)<g'(x_2)$ for any $x_1>x_2>0$ and $x_1+x_2=Q_i+Q_j\leq 1$.

\section{Lemmas}

We give two lemmas to support the proof of Theorem $2$ and Theorem $3$.

\subsection{Lemma $1$}

\begin{Lemma}
\label{lemma-order}
If $Q$ is a Pareto-optimum, then the following conditions are satisfied: if $P_i>P_j$, then $Q_i\geq Q_j$; if $P_i=P_j$, then $Q_i=Q_j$.
\end{Lemma}

If $P_i>P_j$, assume $Q_i<Q_j$, we can construct $Q'$ where $Q'_k=Q_k$ for all $k\neq i,j$ and $Q'_i=Q_j, Q'_j=Q_i$. As such, $V(Q')=V(Q)$ but $U(Q')-U(Q)=(Q_j-Q_i)[f(P_i)-f(P_j)]>0$. This means $Q$ is dominated by $Q'$, which conflicts with the fact that $Q$ is a Pareto-optimum. So $Q_i\geq Q_j$.

If $P_i=P_j$, assume $Q_i\neq Q_j$, and we can further assume $Q_i>Q_j$. Again we construct $Q'$ where $Q'_k=Q_k$ for all $k\neq i,j$ and $Q'_i=Q'_j=\frac{Q_i+Q_j}{2}$. Surely we have $U(Q')=U(Q)$, and $V(Q')-V(Q)=2g(\frac{Q_i+Q_j}{2})-g(Q_i)-g(Q_j)$. Since $g$ is strictly concave, we have $V(Q')-V(Q)>0$, which means $Q$ is dominated by $Q'$. This causes confliction, so $Q_i=Q_j$.

\subsection{Lemma $2$}

\begin{Lemma}
\label{lemma-combine}
Assume $\alpha \in [0,1)$ and $\Psi(Q)=\alpha U(Q)+(1-\alpha)V(Q)$, then the distribution $Q$ that maximize $\Psi(Q)$ satisfies $Q_i=\hat g'^{-1}[w\cdot f(P_i)+b]$, and $w=\frac{\alpha}{\alpha-1}$.
\end{Lemma}

Define the optimization problem as follows:
\begin{equation*}
\begin{aligned}
\min_{Q} -\alpha\cdot U(Q)-(1&-\alpha)V(Q) \\
s.t.\ 1-\sum_{i=1}^N Q_i&=0 \\
\forall i,\ -Q_i&\leq 0
\end{aligned}
\end{equation*}

Again we first check that the prerequisites in KKT are all satisfied. $-U(Q)$ is linear and $-V(Q)$ is convex w.r.t. $Q$; $1-\sum_{i=1}^N Q_i$ is affine w.r.t. $Q$;  since all $Q_i$ can be positive, so the inequalities are all strictly feasible.

The Lagrange function is:
\begin{equation*}
\begin{aligned}
L(Q_i, \lambda, \xi_i) = &-\alpha\sum_{i=1}^N Q_i f(P_i) - (1-\alpha)\sum_{i=1}^N g(Q_i) \\
&+ \lambda (1-\sum_{i=1}^N Q_i) - \sum_{i=1}^N \xi_i Q_i, \quad \xi \geq 0.
\end{aligned}
\end{equation*}
Apply KKT and we get the following conditions for a optimal solution:
\begin{equation*}
\begin{aligned}
\forall i,\ \frac{\partial L}{\partial Q_i} = -\alpha f(P_i)-(1&-\alpha)g'(Q_i)-\lambda-\xi_i = 0, \\
\forall i,\ -\xi_i Q_i &= 0
\end{aligned}
\end{equation*}

For $Q_i\neq 0$, there is $\xi_i=0$, so
\begin{equation*}
Q_i=g'^{-1}[\frac{\alpha}{\alpha-1} f(P_i)+\frac{\lambda}{\alpha-1}];
\end{equation*}
for $Q_i=0$, there is $\xi_i>0$, so
\begin{equation*}
\frac{\alpha}{\alpha-1} f(P_i)+\frac{\lambda}{\alpha-1}>g'(0).
\end{equation*}
Denote $w=\frac{\alpha}{\alpha-1}$ and $b=\frac{\lambda}{\alpha-1}$ and combine the two cases together, we get:
\begin{equation*}
Q_i=\hat g'^{-1}[w\cdot f(P_i)+b], \quad w\leq 0,
\end{equation*}
The above derivation is both sufficient and necessary, so we finished the proof.

\section{Proof of Theorem $2$}
We give the proofs for three conclusions individually.

\subsection{Conclusion $1$}
Here we only consider the case with $U(Q)\neq \max_Q U(Q)$, and the case where $U(Q)=\max_Q U(Q)$ will be incorporated into conclusion 3. We try to find a distribution $Q'$ with the highest diversity while quality is not lower than $Q$. Define a convex optimization problem as follows:
\begin{equation*}
\begin{aligned}
\min_{Q'} -V(Q')& \\
s.t.\ U(Q)-U(Q')&\leq 0 \\
1-\sum_{i=1}^N Q'_i&=0 \\
\forall i,\ -Q'_i&\leq 0
\end{aligned}
\end{equation*}
For $Q$ to be a Pareto-optimum, it's necessary for $Q'=Q$ to be a solution of above problem. Thus we try to solve this problem next.

We first check that the prerequisites in KKT are all satisfied. $-V(Q')$ is convex w.r.t. $Q'$; $1-\sum_{i=1}^N Q'_i$ is affine w.r.t. $Q'$; $U(Q)-U(Q')$ and $-Q'_i$ are convex(linear) w.r.t $Q'$; since all $Q'_i$ can be positive and $U(Q)\neq \max_Q U(Q)$, so the inequalities are all strictly feasible.

The Lagrange function is:
\begin{equation*}
\begin{aligned}
L(Q'_i, \lambda, \eta, \xi_i) = &-\sum_{i=1}^N g(Q'_i) + \lambda (1-\sum_{i=1}^N Q'_i) \\
&+ \eta \sum_{i=1}^N (Q_i-Q'_i)f(P_i) - \sum_{i=1}^N \xi_i Q'_i, \\
&\eta, \xi \geq 0.
\end{aligned}
\end{equation*}
Apply KKT and we get the following conditions for a optimal solution:
\begin{equation*}
\begin{aligned}
\forall i,\ \frac{\partial L}{\partial Q'_i} = -g'(Q'_i)-\lambda-\eta f(P_i)-\xi_i &= 0, \\
\eta[U(Q)-U(Q')] &= 0, \\
\forall i,\ -\xi_i Q'_i &= 0.
\end{aligned}
\end{equation*}
Since we need $Q'=Q$ to be a solution, so
\begin{equation*}
\begin{aligned}
\forall i,\ -g'(Q_i)-\lambda-\eta f(P_i)-\xi_i &= 0, \\
\forall i,\ -\xi_i Q_i &= 0.
\end{aligned}
\end{equation*}
For $Q_i\neq 0$, there is $\xi_i=0$, so $Q_i=g'^{-1}[-\eta f(P_i)-\lambda]$; for $Q_i=0$, there is $\xi_i>0$, so $-\eta f(P_i)-\lambda>g'(0)$. Denote $w=-\eta$ and $b=-\lambda$ and combine the two cases together, we get:
\begin{equation*}
Q_i=\hat g'^{-1}[w\cdot f(P_i)+b], \quad w\leq 0,
\end{equation*}
where
\begin{displaymath}
\hat g'^{-1}(x) = \left\{ \begin{array}{ll}
g'^{-1}(x) & \textrm{if $x< g'(0)$,}\\
0 & \textrm{if $x\geq g'(0)$.}
\end{array} \right.
\end{displaymath}
Now we get a necessary condition for $Q$ to be a Pareto-optimum. To make it sufficient, we still require that for any two distributions satisfying this form, no one could dominate another. This property can be proved by combining conclusion $2$ and $3$.

\subsection{Conclusion $2$}
\label{appendix-twb}
We separate the proof into two parts: (1) $b$ is correspondent to $w$; (2) the monotonicity of $b$ w.r.t. $w$.

\textbf{(1)} The sum of all $Q_i$ should be $1$. Denote
\begin{equation*}
T(w, b) = \sum_{i=1}^N \hat g'^{-1}[w\cdot f(P_i)+b].
\end{equation*}
Since $g'(x)$ is strictly monotonically decreasing, so $T(w, b)$ is monotonically non-increasing w.r.t. $b$. If $T(w, b)>0$, there would be a term which is strictly monotonically decreasing w.r.t. $b$, under which condition $T(w, b)$ is strictly monotonically decreasing w.r.t. $b$. Also, $T(w, b)$ is continuous w.r.t. $b$ since $g'^{-1}$ is continuous. When
\begin{equation*}
b=g'(0)-w\cdot f(\max_i P_i),
\end{equation*}
there is
\begin{equation*}
w\cdot f(P_i)+b\geq w\cdot f(\max_i P_i)+b = g'(0),
\end{equation*}
so $T(w, b)=0$; when
\begin{equation*}
b=g'(\frac{1}{N})-w\cdot f(\min_i P_i),
\end{equation*}
there is
\begin{equation*}
w\cdot f(P_i)+b\leq w\cdot f(\min_i P_i)+b = g'(\frac{1}{N}),
\end{equation*}
so $T(w, b)\geq 1$. From above analysis, the value of $T$ can reach $0$ or be greater than $1$. So combining the monotonicity of $T$, there exists and only one $b$ that satisfies $T(w, b)=1$, leading to a rational distribution.

\textbf{(2)} Define $T(w,b)=\sum_{i=1}^N \hat g'^{-1}[w\cdot f(P_i)+b(w)]$ as above. Since $T(w,b)$ represents the total probability of a distribution, so there should be $T(w,b)\equiv 1$, thus $\frac{\mathrm{d}T}{\mathrm{d}w}=0$.
\begin{equation*}
\begin{aligned}
\frac{\mathrm{d}T}{\mathrm{d}w} = \sum_{i\in S} \frac{f(P_i)+b'(w)}{g''\{g'^{-1}[w\cdot f(P_i)+b(w)]\}},
\end{aligned}
\end{equation*}
where $S=\{i|w\cdot f(P_i)+b(w)<g'(0)\}$.
By the condition $\frac{\mathrm{d}T}{\mathrm{d}w}=0$, we get
\begin{equation*}
b'(w)=-\frac{\sum_{i\in S} \frac{f(P_i)}{g''\{g'^{-1}[w\cdot f(P_i)+b(w)]\}}}{\sum_{i\in S} \frac{1}{g''\{g'^{-1}[w\cdot f(P_i)+b(w)]\}}}.
\end{equation*}
Since $g''(x) < 0$, so if $f(x)<0$ for all $x\in [0,1]$, we can get $b'(w)>0$, thus $b$ is strictly monotonically increasing w.r.t. $w$. Similarly, if $f(x)>0$ for all $x\in [0,1]$, we can get $b'(w)<0$, thus $b$ is strictly monotonically decreasing w.r.t. $w$.

\subsection{Conclusion $3$}
We also separate the proof into two parts: (1) the uniqueness of $Q(w)$; (2) the monotonicity of $U$ and $V$ w.r.t. $w$.

\textbf{(1)} Since $P$ is not uniform, so we can denote $B$, $P_{m_1}$, $P_{m_2}$ as they are in the theorem. According to Lemma \ref{lemma-order}, since $P_{m_1}$ is the largest one, so the corresponding $Q_{m_1}$ is also the largest one, which means
\begin{equation*}
Q_{m_1}=\hat g'^{-1}[w\cdot f(P_{m_1})+b]>0.
\end{equation*}
Thus we get
\begin{equation*}
w\cdot f(P_{m_1})+b < g'(0).
\end{equation*}
At the same time, because we can get $Q_i=Q_{m_1}$ if $P_i=P_{m_1}$, so we can sum up all the largest $Q_i$ and get
\begin{equation*}
M\cdot Q_{m_1}\leq \sum_{i=1}^N Q_i=1,
\end{equation*}
we can get
\begin{equation}
\label{T4-0}
w\cdot f(P_{m_1})+b \geq g'(\frac{1}{M}).
\end{equation}

Consider the case where $w\geq B$, we first prove that $w\cdot f(P_{m_2})+b\leq g'(0)$. Assume
\begin{equation}
\label{T4-1}
w\cdot f(P_{m_2})+b> g'(0),
\end{equation}
then $Q_{m_2}=0$, and there is $Q_i=0$ for any $i$ satisfying $P_i\leq P_{m_2}$. As a result, there should be $Q_i=\frac{1}{M}$ for all $i$ satisfying $P_i=P_{m_1}$, which means
\begin{equation}
\label{T4-2}
w\cdot f(P_{m_1})+b = g'(\frac{1}{M}).
\end{equation}
Subtract Equation \ref{T4-2} by Equation \ref{T4-1}, we get
\begin{equation*}
w\cdot [f(P_{m_1})-f(P_{m_2})] < g'(\frac{1}{M})-g'(0),
\end{equation*}
so
\begin{equation*}
w < \frac{g'(\frac{1}{M})-g'(0)}{f(P_{m_1})-f(P_{m_2})}=B.
\end{equation*}
This contradict with the fact that $w\geq B$. Thus we have $w\cdot f(P_{m_2})+b\leq g'(0)$.

\begin{figure*}[]
\centering
\includegraphics[width=0.4\textwidth]{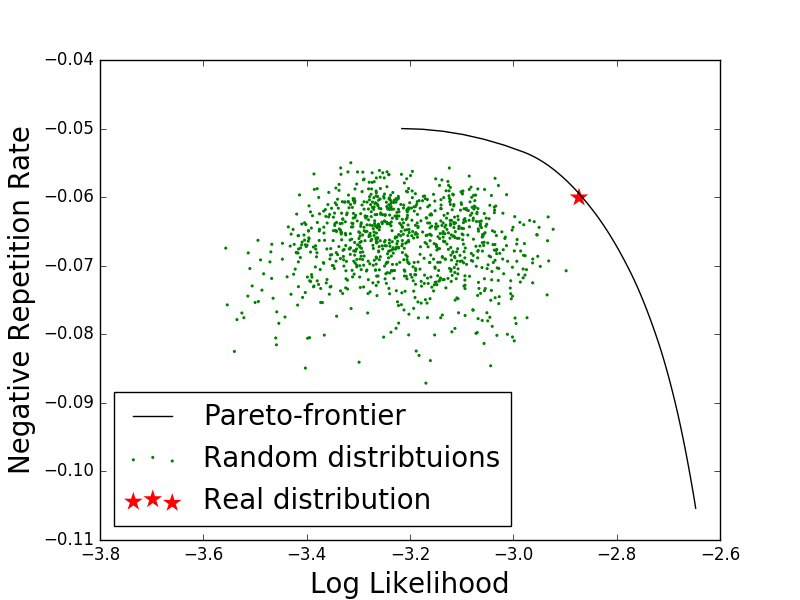}
\includegraphics[width=0.4\textwidth]{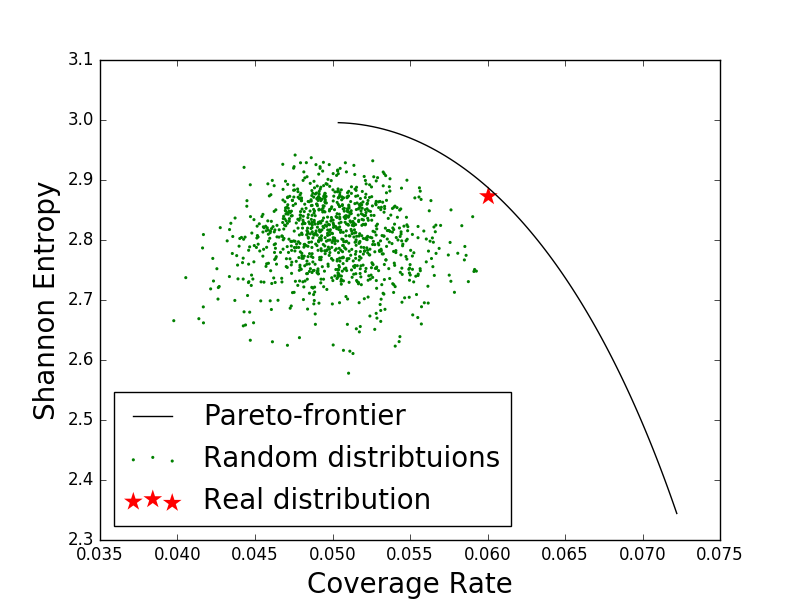}
\caption{Illustration of the Pareto-frontier on a random toy categorical distribution with size $20$. The ground truth distribution is under the frontier curve. \textbf{Left:} Pair LL with NRR. \textbf{Right:} Pair CR with SE.}
\label{figure-mismatch}
\end{figure*}

Combining the above conclusions, for any $w_1,w_2\in [B, 0]$, assume $Q(w_1)=Q(w_2)$, then
\begin{equation*}
\begin{aligned}
w_1\cdot f(P_{m_1})+b_1 &= w_2\cdot f(P_{m_1})+b_2, \\
w_1\cdot f(P_{m_2})+b_1 &= w_2\cdot f(P_{m_2})+b_2.
\end{aligned}
\end{equation*}
As $P_{m_1}\neq P_{m_2}$, so $w_1=w_2$, causing contradiction. Thus we have $Q(w_1)\neq Q(w_2)$.

For any $w\leq B$, assume
\begin{equation}
\label{T4-3}
w\cdot f(P_{m_2})+b < g'(0).
\end{equation}
By subtracting Equation \ref{T4-0} and Equation \ref{T4-3}, we get
\begin{equation*}
w\cdot [f(P_{m_1})-f(P_{m_2})] > g'(\frac{1}{M})-g'(0),
\end{equation*}
so
\begin{equation*}
w > \frac{g'(\frac{1}{M})-g'(0)}{f(P_{m_1})-f(P_{m_2})}=B.
\end{equation*}
This causes contradiction, so the above assumption does not hold. Thus we have $w\cdot f(P_{m_2})+b\geq g'(0)$, which means $Q_{m_2}=0$. Borrowing the proof above, we know that
$Q_i=\frac{1}{M}$ for all $i$ satisfying $P_i=P_{m_1}$. This is a trivial Pareto-optimal case where $U(Q)=\max_Q U(Q)$. Now we know the distribution $Q$ is fixed and does not change as $w$ changes, so for any $w_1,w_2 \leq B$, there is $Q(w_1)=Q(w_2)$.

\textbf{(2)} For the expression of $Q_i$, since $f$ and $g'$ are both continuous and monotonic, so it is easy to know that $Q_i$ is continuous w.r.t. $w$, then $U(Q(w))$ and $V(Q(w))$ are both continuous w.r.t. $w$. We just need to prove the monotonicity.

Assume $B\leq w_1<w_2\leq 0$, the goal is to prove that $U(Q(w_1))>U(Q(w_2))$ and $V(Q(w_1))<V(Q(w_2))$. According to Lemma \ref{lemma-combine}, $w_1$ and $w_2$ have their corresponding $\alpha_1=\frac{w_1}{w_1-1}$ and $\alpha_2=\frac{w_2}{w_2-1}$, and $\alpha_1>\alpha_2$. Since $Q(w)$ is the optimal solution for problem $\alpha U(Q)+(1-\alpha)V(Q)$, and $Q(w_1)$ is different with $Q(w_2)$, so the following inequalities hold:
\begin{equation*}
\begin{aligned}
\alpha_1 U(Q(w_1))+&(1-\alpha_1)V(Q(w_1)) > \\
&\alpha_1 U(Q(w_2))+(1-\alpha_1)V(Q(w_2)), \\
\alpha_2 U(Q(w_1))+&(1-\alpha_2)V(Q(w_1)) < \\
&\alpha_2 U(Q(w_2))+(1-\alpha_2)V(Q(w_2)).
\end{aligned}
\end{equation*}
Subtracting the first equation by the second one, we get
\begin{equation*}
\begin{aligned}
&[(U(Q(w_1))-U(Q(w_2)))-(V(Q(w_1))-V(Q(w_2)))] \\
&\cdot(\alpha_1-\alpha_2) > 0.
\end{aligned}
\end{equation*}
As $\alpha_1>\alpha_2$, so
\begin{equation*}
U(Q(w_1))-U(Q(w_2)) > V(Q(w_1))-V(Q(w_2)).
\end{equation*}
Because $Q(w_1)$ and $Q(w_2)$ are both Pareto-optima, there quality and diversity should satisfy one of the following: $U(Q(w_1))>U(Q(w_2)), V(Q(w_1))<V(Q(w_2))$ or $U(Q(w_1))<U(Q(w_2)), V(Q(w_1))>V(Q(w_2))$. With the derived restriction $U(Q(w_1))-U(Q(w_2)) > V(Q(w_1))-V(Q(w_2))$, we know the first one holds, that is $U(Q(w_1))>U(Q(w_2))$ and $V(Q(w_1))<V(Q(w_2))$.

\begin{figure*}[]
\centering
\includegraphics[width=0.32\textwidth]{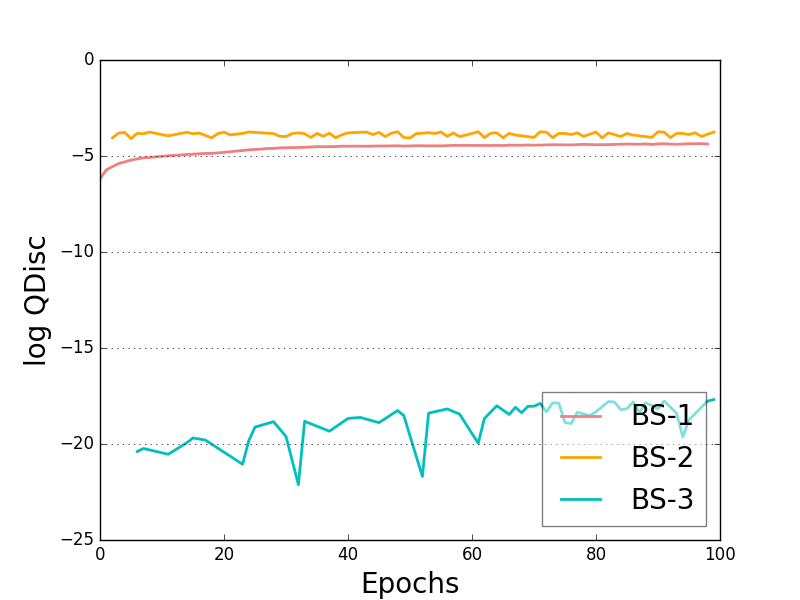}
\includegraphics[width=0.32\textwidth]{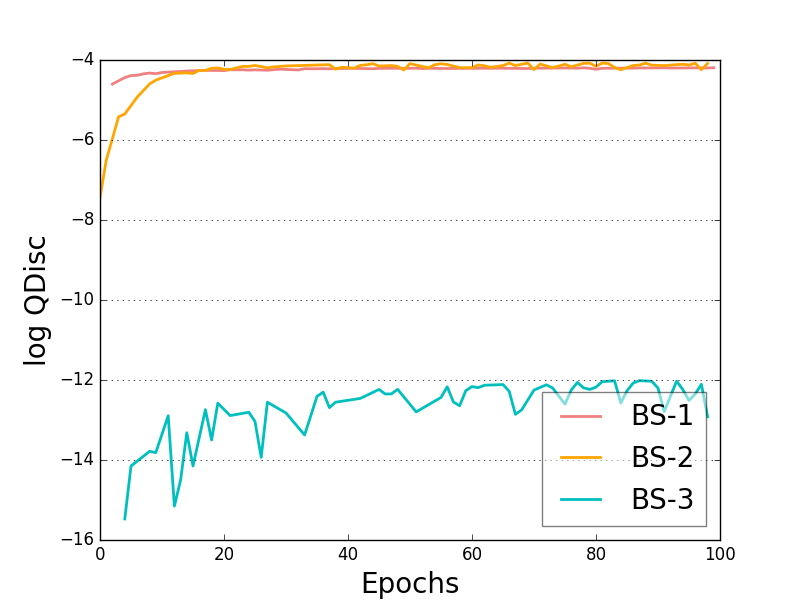}
\includegraphics[width=0.32\textwidth]{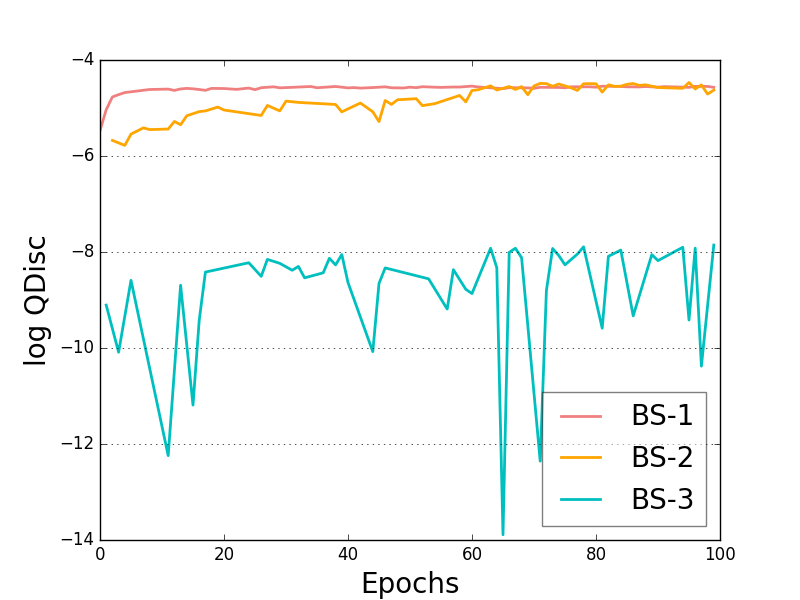}
\caption{The optimization curve of Quality-Discrepancy for BLEU-NSBLEU metric pair on synthetic data with different standard deviations, $\sigma= 0.5, 1.0, 2.0$ from left to right.}
\label{figure-qdisc}
\end{figure*}

\begin{figure*}[]
\centering
\includegraphics[width=0.32\textwidth]{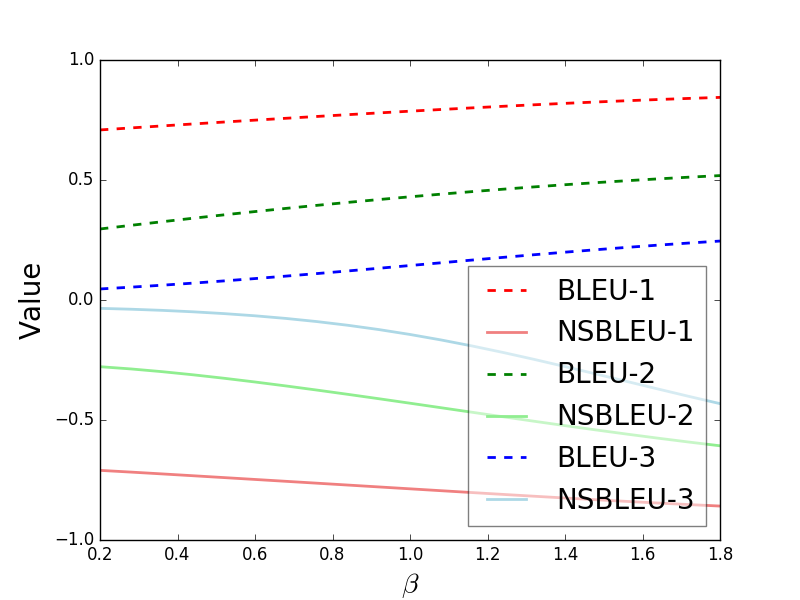}
\includegraphics[width=0.32\textwidth]{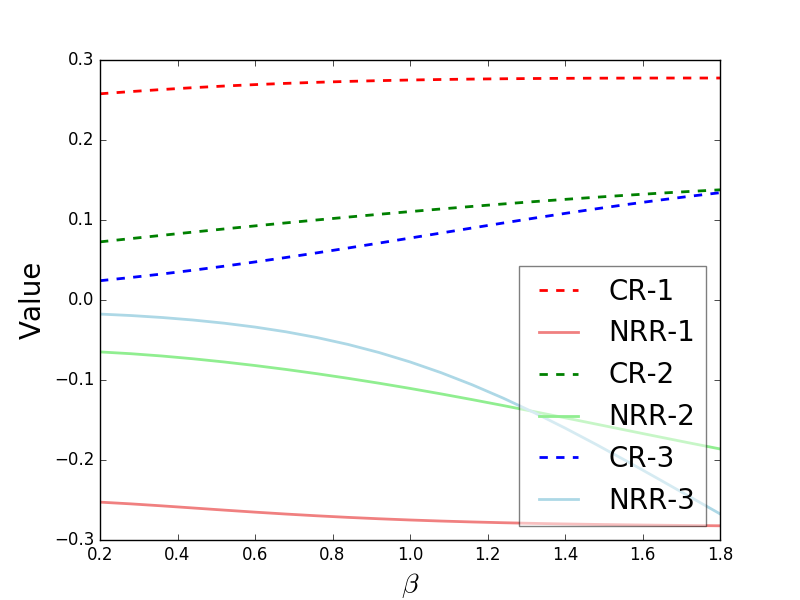}
\includegraphics[width=0.32\textwidth]{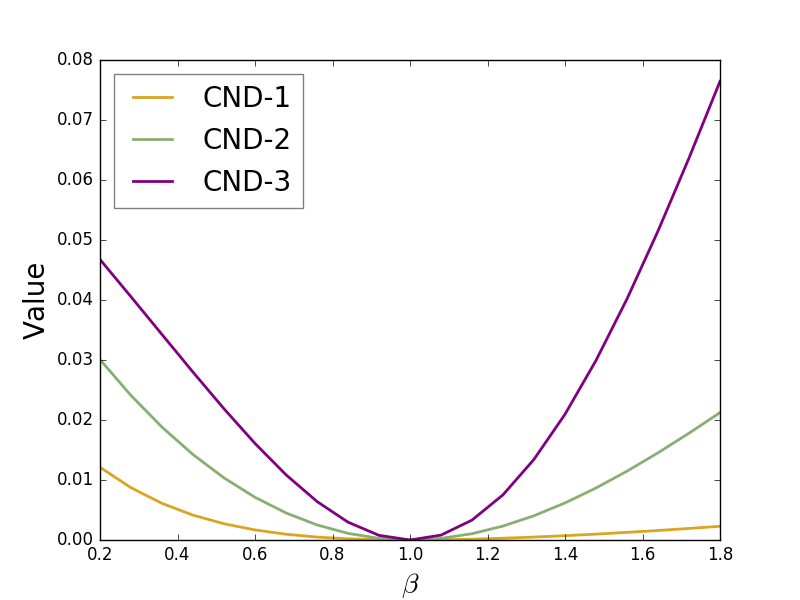}
\caption{Evaluation of BLEU-NSBLEU, CR-NRR, and CND on synthetic data with $\sigma=1.0$. Test models are Pareto-optima parameterized by $\beta$ under LL-SE metric pair.}
\label{figure-trend}
\end{figure*}

\section{Proof of Theorem $3$}

The requirement that $Q=P$ being a Pareto-optimum is equivalent to the following condition: for any $P$, there exist $w_0\leq 0$ and $b_0$ that for any $i$, there is
\begin{equation*}
P_i=\hat g'^{-1}[w_0\cdot f(P_i)+b_0].
\end{equation*}
This means, for any $P_i > 0$, there is $w_0\cdot f(P_i)+b_0=g'(P_i)$.
Since $f$ and $g'$ are both continuous, so
\begin{equation*}
w_0\cdot f(0)+b_0-g'(0) = \lim_{P_i \to 0} w_0\cdot f(P_i)+b_0-g'(P_i) = 0.
\end{equation*}
We can see $w_0\cdot f(P_i)+b_0=g'(P_i)$ is also true for $P_i=0$. By solving this differential equation, we get
\begin{equation*}
g(x)=w_0\int_0^x f(u)\mathrm{d}u+b_0x.
\end{equation*}
Here $b_0$ can be any value because $P_i=g'^{-1}[w_0\cdot f(P_i)+b_0]$ always lead to a plausible distribution $P$.
Under this condition, we know that $Q=P$ is the only distribution that maximize $\Psi(Q)=\alpha U(Q)+(1-\alpha)V(Q)$ where $\alpha=\frac{w_0}{w_0-1}$ according to Lemma \ref{lemma-combine}.
With above conclusions, it is easy to check that $D(P||Q)=\Psi(P)-\Psi(Q)\geq 0$ and $D(P||Q)=0$ if and only if $Q=P$, thus $D(P||Q)$ is a divergence metric.

\section{Pareto-frontier with Mismatched Metrics}
\label{appendix-mismatch}

We show in Figure \ref{figure-mismatch} that the point $Q=P$ is under the Pareto-frontier curve when quality and diversity metrics are not matched, i.e. the condition in Theorem $3$ is not satisfied. We use the same toy dataset, but pair LL with NRR and CR with SE. Note that there is always a gap between the star and the curve, indicating that the real distribution lies on neither of the two Pareto-frontiers.

\section{Additional Information for Experiments}

\subsection{Experiments on Synthetic Data}

\begin{figure}[]
\centering
\includegraphics[width=0.4\textwidth]{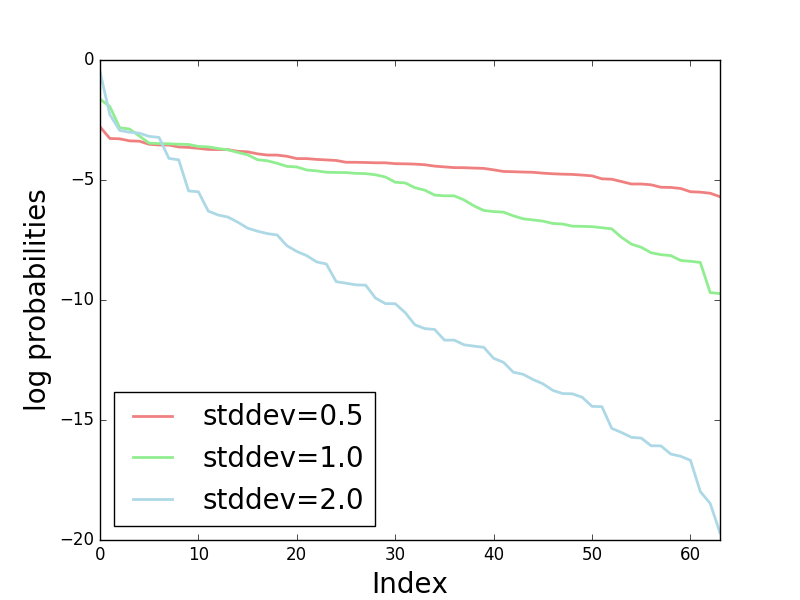}
\caption{The log-probabilities of three synthetic ground truth distributions used in our experiments, shown in descending order.}
\label{figure-toydata}
\end{figure}

The probabilities of synthetic ground truth distributions are shown in Figure \ref{figure-toydata}. We use different standard deviations to get different kind of distributions. Distribution with $\sigma=0.5$ is more flat and of higher entropy, and distribution with $\sigma=2.0$ is more sharp and of lower entropy.

We show the training curve of the optimization process used on synthetic data in Figure \ref{figure-qdisc}. Learning rates are adjusted according to each process, so as to find a best distribution. Points are neglected if $V(Q)\leq V(P)$ or $U(Q)\leq U(P)$, i.e. they fail to dominate the ground truth distribution.

We show the correlation between CR/NRR/CND and quality/diversity/divergence on synthetic data, respectively. We use the well-defined Pareto-frontier under LL-SE in text space as target models, i.e. $Q_i\propto P_i^\beta$. As $\beta$ decreases, the corresponding Pareto-optimum becomes more close to uniform distribution, so that quality decreases and diversity increases according to Theorem $2$, and minimal divergence is taken when $\beta=1$ according to Theorem $3$. We plot the curves of BLEU-NSBLEU, CR-NRR, and CND in Figure \ref{figure-trend}. We can see CR/NRR/CND can properly reflect quality/diversity/divergence, respectively.

\subsection{Experiments on Real Text Data}

For MSCOCO dataset, we remove words with frequency lower than 20, as well as sentences containing them. The vocabulary size is 5,473, and maximum text length is 32. Sentences longer than 32 are also removed, and we get a total number of 530,093 sentences. We randomly sample 50,000 sentences as candidate set, 50,000 sentences as reference set, and another 200,000 sentences for training data of the RNNLM.

For WMT dataset, we use the Europarl-v7 part. We remove words with frequency lower than 400, as well as sentences containing them. The vocabulary size is 6,655, and maximum text length is 50. Sentences longer than 50 or shorter than 20 are also removed, and we get a total number of 475,662 sentences. We again randomly sample 50,000 sentences as candidate set, 50,000 sentences as reference set, and another 200,000 sentences for training data of the RNNLM.

\end{document}